\documentclass{article}

\usepackage{arxiv}

\usepackage[utf8]{inputenc} 
\usepackage[T1]{fontenc}    
\usepackage{hyperref}       
\usepackage{url}            
\usepackage{booktabs}       
\usepackage{amsfonts}       
\usepackage{nicefrac}       
\usepackage{microtype}      
\usepackage{lipsum}		

\usepackage{epsfig}
\usepackage{graphicx}
\usepackage{multirow}
\usepackage{url}

\title{A Study of Human Summaries of Scientific Articles}

\author{
  Odellia Boni \\
  IBM Research AI \\
  \texttt{odelliab@il.ibm.com} \\
  \And
   Guy Feigenblat\\
  IBM Research AI \\
  \texttt{ guyf@il.ibm.com} \\
  \And
    Doron Cohen \\
  IBM Research AI \\
  \texttt{doronc@il.ibm.com} \\
  \And
    Haggai Roitman\\
  IBM Research AI \\
  \texttt{haggai@il.ibm.com}  \\
  \And
   David Konopnicki \\
  IBM Research AI \\
  \texttt{davidko@il.ibm.com}
}

\begin{document}
\maketitle

\begin{abstract}
Researchers and students face an explosion of newly published papers which may be relevant to their work. This led to a trend of sharing human summaries of scientific papers. We analyze the summaries shared in one of these platforms \textit{Shortscience.org}. The goal is to characterize human summaries of scientific papers, and use some of the insights obtained to improve and adapt existing automatic summarization systems to the domain of scientific papers.
\end{abstract}

\section{Introduction}\label{intro}
Recent years have witnessed a boom in the amounts of scientific works being published in various online sources, such as \textit{arXiv.org}, \textit{Google Scholar}, \textit{Microsoft Academic Search}, and \textit{IBM Science Summarizer}  ~\cite{DBLP:conf/emnlp/EreraSFNBRCWMRL19}. 
For example, within less than a decade, the number of yearly submissions to \textit{arXiv} repository has nearly doubled\footnote{\scriptsize{\url{https://arxiv.org/help/stats/2019_by_area/index}}}.

Trying to overcome the information overload, several online sources such as \textit{paper a day}\footnote{\scriptsize{\url{https://medium.com/@sharaf}}}, \textit{The morning Paper}\footnote{\scriptsize{ \url{https://blog.acolyer.org/}}}, \textit{TopBots}\footnote{ \scriptsize{\url{ https://www.topbots.com/most-important-ai-research-papers-2018/}}}, 
\textit{OpenReview.net}  and \textit{ShortScience}~\cite{cohen2017shortscience}, now provide access to human authored summaries of selected works written by both experts and practitioners in their respective communities. Such summaries tend to be long, detailed and contain headlines and figures from the original papers. 

\subsection{Towards automatic summarization}
Scientific papers have a complex structure as well as an intricate content, making their summarization a hard task even for humans. Trying to study the automatic scientific summarization task, several datasets, such as \textit{Scisumm}~\cite{jaidka2016overview}, and \textit{ScisummNet}~\cite{yasunaga2019scisummnet} have been proposed. 
Yet, compared to real human summaries such as the ones nowadays available in the various online sources, existing datasets only focus on automatic generation of relatively short summaries ($150-200$ words) which have an abstract-like structure and are lacking other summarization constructs used by humans such as headlines and figures. Moreover, many of existing summarization methods of scientific papers, rely on citations in order to pinpoint the important parts within scientific papers~\cite{qazvinian2008scientific}. However, for most newly published papers, as the ones usually summarized by humans, citations volume is not large enough to perform a similar analysis.

Trying to fill the gap, we study a dataset for scientific summarization, based on long human summaries authored by \textit{ShortScience.org}\footnote{\scriptsize{\url{https://shortscience.org}}} users. Our goal is to study the characteristics of human scientific summaries and to propose the use of such summaries, that are sometimes published as blogs,  as a potential benchmark for automation in this difficult task.


\section{Dataset}

\textit{ShortScience.org} is an open platform for publishing summaries of scientific papers in the domains of Computer Science (CS), Physics and Biology. 
In this work, we focus on CS publications. The web-site provides minimal instructions on how to write a summary, and therefore, there is a large variation in summary length and structure. To this end, we fetched $561$ summaries from the website associated with $491$ papers. To analyze papers and their summaries, we utilized NLTK for word tokenization and sentence segmentation. We disregard sentences having less than $20$  characters, to minimize effect of parsing errors. The mean summary length is $447$ words and the median is $312$ words.  The average number of sentences per summary is $22$. 

Each summary page links to a publisher website that hosts the source article which we have used to in order to download its PDF version. We have legally fetched articles from the following sources: \textit{Arxiv}
, \textit{NeurIPS}
, \textit{ACL} 
, and \textit{Springer} 
.

We used Science-Parse\footnote{\scriptsize{\url{https://github.com/allenai/science-parse}}} to extract the PDF text of each article. Science-Parse outputs a JSON record for each PDF, which among other fields, contains the title, abstract text, metadata (such as authors and year), and a flat list of the article sections, where each record holds the section's title and text.

\section{Human Summaries Analysis}

\subsection{Summary subjectivity}
Trying to assess to what extent the summaries represent a subjective account of the original scientific work, we explored the expression of opinions by human summarizers. For each summary we extracted all sentences that contained the terms ``i'' or ``my''. Overall out of $561$ summaries we found $130$ summaries that include such sentences. 
To validate the assumption that these terms can be coined with opinion expression and to learn what is the polarity of the opinion (``positive'', ``negative'', or ``neutral''), we conducted a simple evaluation task. We asked five students to read the extracted opinion text from each summary and to indicate if it indeed expresses an opinion and what is its polarity. To aggregate the results across judges, we used a majority vote rule. The analysis shows that out of the $130$ summaries that were tagged as containing opinions, only in $5$ cases this tagging was errorneous. With respect to polarity, surprisingly, most of the summaries were marked as neutral $53\%$, $32\%$ were marked as positive and the rest $15\%$ as negative. This can demonstrate that when humans decide to publicly express their opinion on scientific work they tend to present a positive or balanced view and not to criticize. This can also indicate that people choose to summarize papers they deem valuable.

\subsection{Summary coverage}
Scientific papers are usually structured into several logical categories which address various aspects of the reported research work. 
To assess to what extent human summaries cover such logical aspects of the papers being summarized,  we tried to align each summary sentence to its most probable category in the original paper. To this end, paper sections hierarchy was restored, with sub-sections being merged into their containing high level section. The following high-level sections where identified based on section categories: \textit{Introduction}, \textit{Related work}, \textit{Method}, \textit{Results}, \textit{Experiments}, \textit{Discussion},  \textit{Conclusions}, \textit{Future work} and \textit{Unknown}.  Section sentences inherit their containing section title. Each human summary sentence was then aligned to the paper sentence most similar to it, and was assigned with the category of that sentence. We experimented with three similarity methods: F1 ROUGE-L, average of F1 ROUGE-1, ROUGE-2, and ROUGE-L,  and Cosine similarity over word vectors (GoogleNews-Vectors). 
Overall, $2051$ out of $3421$ article sections were assigned with a category (while the rest were classified as \textit{Unknown}).
Table \ref{tab:roles-table} reports the distribution of summary sentences to logical summary categories. 


\begin{table}[tbh]
\centering
\begin{tabular}{|l|c|c|c|c|}
\hline
\multicolumn{1}{|c|}{\multirow{2}{*}{Category}} & \multicolumn{3}{c|}{\begin{tabular}[c]{@{}c@{}} Category distribution\end{tabular}} \\ \cline{2-4}
\multicolumn{1}{|c|}{} &  \multicolumn{1}{l|}{ROUGE-L} & \begin{tabular}[c]{@{}c@{}}Average ROUGE 1-2-L\end{tabular} & \multicolumn{1}{l|}{Cosine similarity} \\ \hline
Experiments & 27.9\% & 28.0\% & 28.6\% \\
Introduction & 28.3\% & 28.5\% & 28.2\% \\
Method &  14.7\% & 14.0\% & 13.5\% \\
Results &  8.3\% & 8.7\% & 9.4\% \\
Related work &  9.7\% & 9.2\% & 8.3\% \\
Discussion &  5.5\% & 5.1\% & 6.6\% \\
Conclusions &  5.5\% & 6.4\% & 5.3\% \\
Future work &  0.1\% & 0.1\% & 0.1\% \\ \hline
\end{tabular}
\caption{Summary sentence to category distribution}
\label{tab:roles-table}
\end{table}

As depicted in the table, the weights are quite stable when using different similarities, providing further confidence in the correctness of this distribution. Potentially, a summarization algorithm can aim at assigning higher focus to more salient logical sections, reflecting how humans attend different sections in their summary.

%

%


\subsection{Summary style}

\subsubsection{Figures inclusion}: Figures are used in scientific papers to illustrate an architecture or a flow, to exemplify a statement or report results in a graphical way. Some human summaries in the dataset include figures from the original paper including image captures of equations or tables. 
Overall, about $31\%$ of the summaries include at least one such figure, with an average of $2$ figures per summary. 
This suggests that, in some cases, human summarizers would assume that the best way to explain an idea and deliver it in a more understandable form is to utilize visual aids along with the accompanied summary text. This demonstrates the need to consider multi-modal summarization, which to the best of our knowledge, has not been research yet in this domain.

\subsubsection{Summary Itemization:} Almost half of the summaries in the dataset utilized some form of structuring using itemization (i.e., bullets or numbering), with an average of $15$ items per summary. We further measured the amount of text associated with each item by counting the number of sentences between each two consecutive items. The average size of an item is $2$ sentences. This conforms with the typical usage of such a structure: each item most probably conveys a single concise fact. This is also supported by a strong positive correlation $(r=0.693)$ between number of items in a summary and summary length measured in sentences. 

\subsubsection{Headlines:} About $35\%$ of the summaries contain lines that start with ``\#'', which act as summary ``headlines''.  Out of $85$ summary authors $35$ used headlines in their summaries. Such headlines are commonly used by human summarizers for dividing their summary into small logical parts. In order to determine which subjects prevail in creating summary subdivisions, we analyzed summary headlines text. Table \ref{tab:MostFrequentUnigramsAndBigramsInSummariesHeadlines} contains the most frequent unigrams and bigrams in summary headlines (after tokenization, stemming and stopwords removal). From the table it seems that many of the subdivisions either refer to specific logical aspect of the paper (as hinted by words like ``result'', ``model'', ``dataset'' ) or convey summarizer's take on the paper (``my two cents'', ``key points'').
\begin{table}[tbh]
	\centering
	
		\begin{tabular}{|c|c||c|c|}
		\hline
		Unigrams    & Frequency     & Bigrams       & Frequency \\
		\hline
		result    & 63           & (key,point)   & 34 \\
		model     & 44           & (my, two)     & 17 \\
		note      & 44           & (two, cent)  & 17 \\
		key       & 43           & (weak, note)  & 10 \\
		point     & 34           & (rough, note) & 9 \\
    architectur &  30        & (problem, address)& 8 \\
    introduct & 28           & (key, takeaway) & 8 \\
    dataset   & 28           & (see, also)   &  8 \\
		\hline
		\end{tabular} 
		
	\caption{Frequent unigrams and bigrams in headlines}
	\label{tab:MostFrequentUnigramsAndBigramsInSummariesHeadlines}
\end{table}

\bibliographystyle{unsrt}
\bibliography{shortscience}

\begin{thebibliography}{1}

\bibitem{DBLP:conf/emnlp/EreraSFNBRCWMRL19}
Shai Erera, Michal Shmueli{-}Scheuer, Guy Feigenblat, Ora~Peled Nakash, Odellia
  Boni, Haggai Roitman, Doron Cohen, Bar Weiner, Yosi Mass, Or~Rivlin, Guy Lev,
  Achiya Jerbi, Jonathan Herzig, Yufang Hou, Charles Jochim, Martin Gleize,
  Francesca Bonin, and David Konopnicki.
\newblock A summarization system for scientific documents.
\newblock In {\em Proceedings of {EMNLP-IJCNLP} 2019, Hong Kong, China}, pages
  211--216, 2019.

\bibitem{cohen2017shortscience}
Joseph~Paul Cohen and Henry~Z Lo.
\newblock Shortscience. org-reproducing intuition.
\newblock {\em arXiv preprint arXiv:1707.06684}, 2017.

\bibitem{jaidka2016overview}
Kokil Jaidka, Muthu~Kumar Chandrasekaran, Sajal Rustagi, and Min-Yen Kan.
\newblock Overview of the cl-scisumm 2016 shared task.
\newblock In {\em Proceedings of the Joint Workshop on Bibliometric-enhanced
  Information Retrieval and Natural Language Processing for Digital Libraries
  (BIRNDL)}, pages 93--102, 2016.

\bibitem{yasunaga2019scisummnet}
Michihiro Yasunaga, Jungo Kasai, Rui Zhang, Alexander R Fabbri Irene~Li Dan,
  and Friedman Dragomir~R Radev.
\newblock Scisummnet: A large annotated corpus and content-impact models for
  scientific paper summarization with citation networks.
\newblock 2019.

\bibitem{qazvinian2008scientific}
Vahed Qazvinian and Dragomir~R Radev.
\newblock Scientific paper summarization using citation summary networks.
\newblock In {\em Proceedings of the 22nd International Conference on
  Computational Linguistics-Volume 1}, pages 689--696. Association for
  Computational Linguistics, 2008.

\end{thebibliography}

\end{document}